# An ε Hierarchical Fuzzy Twin Support Vector Regression

Arindam Chaudhuri

*Abstract*—The research presents ε-hierarchical fuzzy twin support vector regression (ε-HFTSVR) based on ε-fuzzy twin support vector regression (ε-FTSVR) and ε-twin support vector regression (ε-TSVR). ε-FTSVR is achieved by incorporating trapezoidal fuzzy numbers to ε-TSVR which takes care of uncertainty existing in forecasting problems. ε-FTSVR determines a pair of ε-insensitive proximal functions by solving two related quadratic programming problems. The structural risk minimization principle is implemented by introducing regularization term in primal problems of ε-FTSVR. This yields dual stable positive definite problems which improves regression performance. ε-FTSVR is then reformulated as ε-HFTSVR consisting of a set of hierarchical layers each containing ε-FTSVR. Experimental results on both synthetic and real datasets reveal that ε-HFTSVR has remarkable generalization performance with minimum training time.

## I. Introduction

SUPPORT vector machines (SVMs) are powerful tools for pattern classification and regression [1]. They have been successfully applied to several real world problems [2]. There exist some classical methods [3] where decision surface is found by maximizing the margin between parallel hyperplanes. Recently some nonparallel hyperplane classifiers such as twin support vector regression (TSVR) [4] are developed where two nonparallel proximal hyperplanes are used such that each hyperplane is closest to one class and farther than other class. TSVR solves two smaller sized quadratic programming problems and is faster than classical approaches. It is excellent at dealing with cross planes dataset. Other methods like ε-support vector regression (ε-SVR) [5] finds a linear function such that more training samples locate in ε-insensitive tube and function is as flat as possible leading to structural risk minimization principle. Another commonly used regressor viz. ε-twin support vector regression (ε-TSVR) [6] behaves like TSVR but it minimizes structural risk by adding regularization term through two functions that are as flat as possible. The dual problems are derived without any extra assumption and need not be modified any more. The experiments have shown that ε-TSVR is faster and has better generalization. Based on this motivation, we first illustrate ε-fuzzy twin support vector regression (ε-FTSVR) which is an extension of ε-TSVR. ε-FTSVR is then remodeled as ε-hierarchical fuzzy twin support vector regression (ε-HFTSVR) which consists of set of hierarchical layers each containing ε-FTSVR with gaussian kernel at given scale. On increasing scale layer by layer details are incorporated inside regression function. It adapts local scale to data keeping number of support vectors and comparable configuration time. The approach is based on interleaving regression estimate with pruning activity. ε-HFTSVR is applied to noisy synthetic and real datasets. It denoises original data obtaining an effective reconstruction of better quality. The major contributions of this work include: (a) fuzzification [7] of ε-TSVR leading to ε-FTSVR (b) hierarchical formulation of ε-FTSVR for noisy mislabeled samples to bring robustness in classification results. This paper is presented as follows. In section II, we introduce ε-TSVR. This is followed by ε-FTSVR and ε-HFTSVR in sections III and IV respectively. In next section experimental results are highlighted. Finally, in section VI conclusions are given.

## II. ε-Twin Support Vector Regression

The ε-TSVR is formalized based on TSVR [4] and ε-SVR [5]. The ε-TSVR concentrates on two ε-insensitive proximal linear functions:

$$h_1(x) = w_1^T x + b_1 \quad (1) \qquad h_2(x) = w_2^T x + b_2 \quad (2)$$

The empirical risks are measured by:

$$R_{em}^{\varepsilon_1}[h_1] = \sum_{i=1}^{m} \max\{0, (y_i - h_1(x_i))^2\} + p_1 \sum_{i=1}^{m} \max\{0, -(y_i - h_1(x_i) + \varepsilon_1)\} \quad (3)$$

$$R_{em}^{\varepsilon_2}[h_2] = \sum_{i=1}^{m} \max\{0, (h_2(x_i) - y_i)^2\} + p_2 \sum_{i=1}^{m} \max\{0, -(h_2(x_i) - y_i + \varepsilon_2)\} \quad (4)$$

Here, $p_1 > 0$, $p_2 > 0$ and $\sum_{i=1}^{m} \max\{0, -(y_i - h_1(x_i) + \varepsilon_1)\}$ and $\sum_{i=1}^{m} \max\{0, -(h_2(x_i) - y_i + \varepsilon_2)\}$ are the one side ε-insensitive loss function [5]. By introducing regularization terms $\frac{1}{2}(w_1^T w_1 + b_1^2)$ and $\frac{1}{2}(w_2^T w_2 + b_2^2)$, slack variables $\xi, \xi^*, \eta$ and $\eta^*$, the primal problems are expressed as:

Arindam Chaudhuri is with Samsung Research & Development Institute Delhi, Noida – 201304 INDIA (corresponding author phone: +919871466996; e-mail: ar.chaudhuri@samsung.com).

$$\min_{w_1,b_1,\xi,\xi^*} \frac{1}{2}p_3(w_1^T w_1 + b_1^2) + \frac{1}{2}\xi^{*T}\xi^* + p_1 e^T \xi$$
subject to: $Y - (Aw_1 + eb_1) = \xi^*$ (5)
$Y - (Aw_1 + eb_1) \geq -\varepsilon_1 e - \xi, \xi \geq 0$

and

$$\min_{w_2,b_1,\xi,\xi^*} \frac{1}{2}p_4(w_2^T w_2 + b_2^2) + \frac{1}{2}\eta^{*T}\eta^* + p_2 e^T \eta$$
subject to: $(Aw_2 + eb_2) - Y = \eta^*$ (6)
$(Aw_2 + eb_2) - Y \geq -\varepsilon_2 e - \eta, \eta \geq 0$

Here $p_3 > 0, p_4 > 0, \varepsilon_1 > 0$ and $\varepsilon_2 > 0$. The solutions of equations (5) and (6) are obtained by deriving their dual problems. The Lagrangian of equation (5) is given by:

$$L(w_1,b_1,\xi,\alpha,\beta) = \frac{1}{2}\big(Y - (Aw_1 + eb_1)\big)^T \big(Y - (Aw_1 + eb_1)\big)$$
$$+ \frac{1}{2}p_3(\|w_1\|^2 + b_1^2) + p_1 e^T \xi - \beta^T \xi$$
$$- \alpha^T(Y - (Aw_1 + eb_1) + \varepsilon_1 e + \xi) \quad (7)$$

Here $\alpha = (\alpha_1, \ldots, \alpha_m)$ and $\beta = (\beta_1, \ldots, \beta_m)$ are vectors of Lagrange multipliers. The Karush Kuhn Tucker condition for $w_1, b_1, \xi, \alpha$ and $\beta$ are given by:

$-A^T(Y - Aw_1 - eb_1) + p_3 w_1 + A^T \alpha = 0$ (8)
$-e^T(Y - Aw_1 + eb_1) + p_3 b_1 + e^T \alpha = 0$ (9)
$p_1 e - \beta - \alpha = 0$ (10)
$Y - (Aw_1 + eb_1) \geq -\varepsilon_1 e - \xi, \xi \geq 0$ (11)
$\alpha^T(Y - (Aw_1 + eb_1) + \varepsilon_1 e + \xi) = 0, \beta^T \xi = 0$ (12)
$\alpha \geq 0, \beta \geq 0$ (13)
Since $\beta \geq 0$ we have: $0 \leq \alpha \leq p_1 e$ (14)

The equations (8)-(9) imply that:

$$-\begin{bmatrix} A^T \\ e^T \end{bmatrix} Y + \left(\begin{bmatrix} A^T \\ e^T \end{bmatrix}[A \quad e] + p_3 I\right)\begin{bmatrix} w_1 \\ b_1 \end{bmatrix} + \begin{bmatrix} A^T \\ e^T \end{bmatrix}\alpha = 0 \quad (15)$$

Assuming $J = [A \quad e], v_1 = [w_1 \quad b_1]^T$ equation (15) is rewritten as:
$$v_1 = (J^T J + p_3 I)^{-1} J^T (Y - \alpha) \quad (16)$$

Then substituting equation (16) into Lagrangian and using Karush Kuhn Tucker conditions the dual problem is:
$$\max_\alpha -\frac{1}{2}\alpha^T J(J^T J + p_3 I)^{-1} J^T \alpha^T + Y^T J(J^T J + p_3 I)^{-1} J^T \alpha$$
$$- (e^T \varepsilon_1 + Y^T)\alpha$$
subject to: $0 \leq \alpha \leq p_1 e$ (17)

In equation (17) adjusting $p_3$ improves classification accuracy. In similar manner dual of equation (6) is obtained. Once solutions $(w_1, b_1)$ and $(w_2, b_2)$ of equations (5) and (6) are obtained from solutions of equation (17) and its dual, proximal functions $h_1(x)$ and $h_2(x)$ are achieved. The estimated regressor considered as approximation function is:
$$h(x) = \frac{1}{2}\big(h_1(x) + h_2(x)\big)$$
$$h(x) = \frac{1}{2}(w_1 + w_2)^T x + \frac{1}{2}(b_1 + b_2) \quad (18)$$

III. ε- FUZZY TWIN SUPPORT VECTOR REGRESSION

In order to deal with problems of finite samples and uncertain data in existing in many forecasting situations, input variables and output function $h(\bar{x})$ are described as crisp numbers by fuzzy comprehensive evaluation. To represent fuzzy degree of input variables, trapezoidal fuzzy membership function is adopted [7]. Suppose fuzzy training sample set $\{\bar{x}_i, y_i\}_{i=1}^m$ with $\bar{x}_i = (s_{x_i}, \Delta s_{x_i}, \overline{\Delta s}_{x_i}, \overline{\overline{\Delta s}}_{x_i}) \in P(R)^d$, $y_i = (s_{y_i}, \Delta s_{y_i}, \overline{\Delta s}_{y_i}, \overline{\overline{\Delta s}}_{y_i}) \in P(R)$ and $P(R)^d$ is $d$ dimensional vector set. In light of ε-FTSVR regression coefficients $P(R)$ is estimated by following constrained optimization problems:

$$\min_{\bar{w}_1,b_1,\xi,\xi^*} \frac{1}{2}p_3(\bar{w}_1^T \bar{w}_1 + b_1^2) + \frac{1}{2}\xi^{*T}\xi^* + p_1 e^T \xi$$
subject to: $Y - (A\bar{w}_1 + eb_1) = \xi^*$ (19)
$Y - (A\bar{w}_1 + eb_1) \geq -\varepsilon_1 e - \xi, \xi \geq 0$

and

$$\min_{w_2,b_1,\xi,\xi^*} \frac{1}{2}p_4(\bar{w}_2^T \bar{w}_2 + b_2^2) + \frac{1}{2}\eta^{*T}\eta^* + p_2 e^T \eta$$
subject to: $(A\bar{w}_2 + eb_2) - Y = \eta^*$ (20)
$(A\bar{w}_2 + eb_2) - Y \geq -\varepsilon_2 e - \eta, \eta \geq 0$

The terms $p_3 > 0, p_4 > 0, \varepsilon_1 > 0$ and $\varepsilon_2 > 0$ are crisp numbers. The Lagrangian of equation (19) is given by:

$$L(\bar{w}_1,b_1,\xi,\alpha,\beta) = \frac{1}{2}\big(Y - (A\bar{w}_1 + eb_1)\big)^T\big(Y - (A\bar{w}_1 + eb_1)\big)$$
$$+ \frac{1}{2}p_3(\|\bar{w}_1\|^2 + b_1^2) + p_1 e^T \xi - \beta^T \xi$$
$$- \alpha^T(Y - (A\bar{w}_1 + eb_1) + \varepsilon_1 e + \xi) \quad (21)$$

The Lagrange multipliers are $\alpha = (\alpha_1, \ldots, \alpha_m)$ and $\beta = (\beta_1, \ldots, \beta_m)$. The Karush Kuhn Tucker condition for $\bar{w}_1, b_1, \xi, \alpha$ and $\beta$ are given by:

$-A^T(Y - A\bar{w}_1 - eb_1) + p_3 \bar{w}_1 + A^T \alpha = 0$ (22)
$-e^T(Y - A\bar{w}_1 + eb_1) + p_3 b_1 + e^T \alpha = 0$ (23)
$p_1 e - \beta - \alpha = 0$ (24)
$Y - (A\bar{w}_1 + eb_1) \geq -\varepsilon_1 e - \xi, \xi \geq 0$ (25)
$\alpha^T(Y - (A\bar{w}_1 + eb_1) + \varepsilon_1 e + \xi) = 0, \beta^T \xi = 0$ (26)
$\alpha \geq 0, \beta \geq 0$ (27)
Since $\beta \geq 0$ we have: $0 \leq \alpha \leq p_1 e$ (28)

The equations (22)-(23) imply that:
$$-\begin{bmatrix} A^T \\ e^T \end{bmatrix} Y + \left(\begin{bmatrix} A^T \\ e^T \end{bmatrix}[A \quad e] + p_3 I\right)\begin{bmatrix} \bar{w}_1 \\ b_1 \end{bmatrix} + \begin{bmatrix} A^T \\ e^T \end{bmatrix}\alpha = 0 \quad (29)$$

Assuming $J = [A \quad e], v_1 = [\bar{w}_1 \quad b_1]^T$ equation (29) is rewritten as:
$$\bar{v}_1 = (J^T J + p_3 I)^{-1} J^T (Y - \alpha) \quad (30)$$

Then substituting equation (30) into Lagrangian and using Karush Kuhn Tucker conditions the dual problem is:
$$\max_\alpha -\frac{1}{2}\alpha^T J(J^T J + p_3 I)^{-1} J^T \alpha^T + Y^T J(J^T J + p_3 I)^{-1} J^T \alpha$$
$$- (e^T \varepsilon_1 + Y^T)\alpha$$
subject to: $0 \leq \alpha \leq p_1 e$ (31)

The approximation function is:
$$h(\bar{x}) = \frac{1}{2}\big(h_1(\bar{x}) + h_2(\bar{x})\big)$$
$$h(\bar{x}) = \frac{1}{2}(\bar{w}_1 + \bar{w}_2)^T \cdot \bar{x} + \frac{1}{2}(b_1 + b_2) \quad (32)$$

In equation (32) $\bar{w}_1 = (\bar{w}_{11}, \ldots, \bar{w}_{1d})$ and $\bar{w}_2 = (\bar{w}_{21}, \ldots, \bar{w}_{2d})$ such that $|\bar{w}_1| = (|\bar{w}_{11}|, \ldots, |\bar{w}_{1d}|)$ and $|\bar{w}_2| = $

$(|\overline{w}_{21}|,..,|\overline{w}_{2d}|))$. The inner product of $\overline{w}_1$ and $\overline{x}$ is $\overline{w}_1 \cdot \overline{x}$. In $P(R)$, $h(\overline{x})$ can be written as:

$$h(\overline{x}) = \frac{1}{2}\left((\overline{w}_1 + \overline{w}_2) \cdot s_x + (b_1 + b_2)\rho(\Delta s_x)\right) \quad (33)$$

In equation (33) $\rho(\Delta s_x) = |(\overline{w}_1 + \overline{w}_2) \cdot \Delta s_x|$ with $\overline{w}_1, \overline{w}_2, s_x, \Delta s_x \in R^d$ and $b_1, b_2 \in R$.

## IV. ε- HIERARCHICAL FUZZY TWIN SUPPORT VECTOR REGRESSION

Based on ε-FTSVR, ε-HFTSVR is formulated here. ε-HFTSVR is constituted into a pool of $V$ layers each comprising of single kernel ε-FTSVR $\{m_v(\circ)\}$ by suitable scale. The different layers are placed in hierarchy having scale determined by parameter $\tau_v$ which increases when layer number decreases $(\tau_v \leq \tau_{v+1})$. The output of ε-HFTSVR is:

$$k(\overline{x}) = \sum_{v=1}^{V} m_v(\overline{x}; \tau_v) \quad (34)$$

ε-HFTSVR configuration proceeds by adding and configuring one layer at a time. It initiates from layer featuring smallest scale to that featuring largest one. The first layer is trained such that distance between regression curve produced by first layer itself and data is minimized. It plays a significant role in its success. It is trained heuristically so that number of used layers reduces. All other layers are trained to approximate the residual. The residual for each layer is:

$$rm_v(\overline{x}_i) = rm_{v-1}(\overline{x}_i) - m_v(\overline{x}_i) \quad (35)$$

The $v^{th}$ layer is configured with training set $TS_v = \{(\overline{x}_1, rm_{v-1}(\overline{x}_1)), \ldots, (\overline{x}_n, rm_{v-1}(\overline{x}_n))\}$. The value of scale parameter of first layer $\tau_1$ is proportional to input domain's size. The parameter $\tau$ is decreased arbitrarily. The most preferred value of $\tau$ for each layer is $\tau_{v+1} = \tau_v/n; n \geq 2$ producing satisfactory results. On decreasing $\tau$ slowly accuracy of solution improves but number of layers and number of support vectors increases. New layers are added during training until stopping criterion is satisfied. The two other parameters are defined for each layer: (a) $B_v$ is tradeoff between regression error and smoothness of solution and (b) $\epsilon$ which controls amplitude of $\epsilon$-insensitivity tube around solution itself. The value of $B$ is usually set experimentally by trial and error. Here, $B_v$ is chosen for each layer as $S$ times variance of residuals used to configure the $v^{th}$ layer as:

$$B_v = Svar(rm_{v-1}(\overline{x}_i)) \quad (36)$$

In equation (36) $B_v$ assumes value taken by Lagrange multipliers associated to support vectors of $v^{th}$ layer which represents maximum weight associated to each kernel. For input space regions where Gaussians associated to support vectors have no significant overlap. This depends both on Gaussian scale parameter and data density. The value of $B_v$ is approximately maximum value that can be assumed by regression function in those regions as Gaussian kernel is 1. For this reason $B_v$ is large enough to allow regression curve reaching maximum or minimum value of data points inside whole input domain. However, a larger $B_v$ favors overfitting. The experimental results on different datasets suggest that $S$ lies in interval $(0, 5]$ which represents a tradeoff. Similar to ε-FTSVR, parameter $\epsilon$ cannot be determined from dataset; rather $\epsilon$ is set proportional to accuracy required for regression. Experiments show that in ε-HFTSVR, layers with larger $\tau$ have number of support vectors similar to layers with smaller $\tau$. There appears some contradiction as fewer units are required to realize a reconstruction at larger scale. Hence, in first layer where ε-HFTSVR output has low frequency content many data points lie far from curve and are still selected as support vectors. This leads to high number of support vectors. To avoid this after each layer has been configured, a pruning step is carried out to reduce number of support vectors. The cost function is then minimized a second time considering only reduced training set to obtain final approximation for each current layer. To reduce number of support vectors it is noticed that distance of training point from regression curve measures suitability of current curve to describe information conveyed. In this sense, points too distant from regression curve cannot be explained by curve. They can be regarded as outliers. For these reasons, acceptable approximation of regression curve is obtained using only those points that lie close to curve. This has been confirmed experimentally. It is observed that quality of regression at given scale does not degrade significantly if regression is computed considering only points close to $\epsilon$-tube. The closeness of point to $\epsilon$-tube can be assessed only after computation of regression itself considering all training points. In second pass, regression is computed again considering only points close to $\epsilon$-tube. Consider $v^{th}$ layer and regression computed for layer $m_v(\overline{x})$ using complete training set $TS_v$. Let us define $TS'_v$ set constituting only of those support vectors that lie on border of $\epsilon$-tube and those whose distance from $m_v(\overline{x}) < \epsilon/n$ as:

$$TS'_v = \left\{(\overline{x}_i, rm_{v-1}(\overline{x}_i)) \Big| ||rm_v(\overline{x}_i)| - \epsilon| < tp \lor |rm_v(\overline{x}_i)| < \frac{\epsilon}{n}\right\} \quad (37)$$

In equation (37) $tp$ is tolerance parameter that determines thickness of $\epsilon$-tube margin. The configuration phase of each layer is structured in two sequential steps: (a) first provides regression curve $m_v$ considering all training points and (b) second $m'_v$ realizes an efficient regression curve by considering only selected subset of points. To cope with diminished point density in $TS'_v$, value of parameter $B_v$ is increased proportionally in second optimization step as:

$$B'_v = B_v \frac{|TS_v|}{|TS'_v|} = Svar(rm_{v-1}(\overline{x}_i)) \frac{|TS_v|}{|TS'_v|} \quad (38)$$

## V. EXPERIMENTAL RESULTS

In this section, some experiments are performed to demonstrate performance of ε-HFTSVR compared with ε-FTSVR and ε-TSVR on synthetic and real datasets. All methods are implemented in Matlab 8.2 on PC having Intel P4 processor with 2.9 GHz and 1 GB RAM with 512 KB cache. The values of parameters are obtained through searching in range $[2^{-9}, 2^9]$ by tuning a set comprising of random 20 % of dataset. In experiments, we set $p_1 = p_2, p_3 = p_4$ and $\varepsilon_1 = \varepsilon_2$ to reduce computational complexity of parameter selection. In order to assess the performance of methods, evaluation criteria used are: (a) SSE (b) NMSE (c) $R^2$ and (d) MAPE.

## A. Synthetic Datasets

The synthetic datasets are taken from [8]. Considering function $y = x^{2/3}$ training samples are distorted by Gaussian noise with 0 mean and 0.2 standard deviation so that:

$$y_i = x_i^{2/3} + \xi_i, x \sim U[-2,2], \xi_i \sim N(0, 0.2^2) \quad (39)$$

In equation (39) $U[a,b]$ and $N(\bar{a}, \bar{b}^2)$ represents uniform and gaussian random variable respectively. To avoid biased comparisons 10 independent groups of noisy samples are generated consisting of 200 training and 200 none noise test samples. It has been observed that ε-HFTSVR achieves best approximation. The results of performance criteria are given (see Table I). ε-HFTSVR derives smallest SSE, NMSE and largest $R^2$ among all methods. This indicates statistical information in training dataset is well explained by ε-HFTSVR with small regression errors. It is observed that ε-HFTSVR is fastest learning method improving training speed. Another datasets are generated by sine function distorted by gaussian noise with 0 mean and 0.2 standard deviation so that:

$$y_i = \frac{sin(x_i)}{x_i} + \xi_i, x \sim U[-4\pi, 4\pi], \xi_i \sim N(0, 0.2^2) \quad (40)$$

The dataset consists of 272 training samples and 526 test samples. The results show the superiority of ε-HFTSVR is demonstrated (see Table I).

TABLE I
COMPARATIVE RESULTS OF ε-HFTSVR WITH OTHER REGRESSOR ON SYNTHETIC DATASETS

| Dataset | Regressor | SSE | NMSE | $R^2$ | CPU(sec) |
|---|---|---|---|---|---|
| Eq (41) | ε-HFTSVR | 0.4252 | 0.0096 | 0.9996 | 0.0025 |
|  | ε-FTSVR | 0.4866 | 0.0112 | 0.9992 | 0.0062 |
|  | ε-TSVR | 0.5250 | 0.0152 | 0.9988 | 0.0080 |
| Eq (42) | ε-HFTSVR | 0.7786 | 0.0069 | 0.9942 | 0.0369 |
|  | ε-FTSVR | 0.9999 | 0.0102 | 0.9925 | 0.0525 |
|  | ε-TSVR | 1.1016 | 0.0203 | 0.9911 | 0.0730 |

## B. Real Datasets

For further evaluation the experimental are also performed on UCI datasets such as Servo and Auto Price [9] in terms of NMSE, $R^2$, MAPE and CPU time. The given results further confirm the superiority of ε-HFTSVR (see Table II). The best parameters selected by ε-HFTSVR and ε-FTSVR on above UCI datasets are given (see Table III). It is observed that the values of $p_3$ and $p_4$ vary and usually do not consider smaller value in ε-HFTSVR. This implies that regularization in terms of ε-HFTSVR is significant.

## VI. CONCLUSION

In this work we propose a novel regressor ε-HFTSVR by minimizing structural risk. The motivation towards developing the regressor is attributed towards ε-FTSVR which is obtained by applying trapezoidal fuzzy numbers to ε-TSVR. ε-FTSVR takes care of uncertainty existing in forecasting problems. It determines a pair of ε-insensitive proximal functions by solving two related SVM type problems. The problem is solved by introducing regularization term in primal problems of ε-FTSVR and handling the dual alternative. This improves overall regression performance. Then ε-HFTSVR is formulated as set of hierarchical layers each containing ε-FTSVR. Experimental results on both synthetic and real datasets reveal that ε-HFTSVR has superior compared to other regressor. However, suitable parameter selection of ε-HFTSVR remains a practical problem towards future research.

TABLE II
COMPARATIVE RESULTS OF ε-HFTSVR WITH OTHER REGRESSOR ON UCI DATASETS

| Dataset | Regressor | NMSE | $R^2$ | MAPE | CPU(sec) |
|---|---|---|---|---|---|
| Servo | ε-HFTSVR | 0.186 ± 0.096 | 0.986 ± 0.169 | 0.315 ± 0.136 | 0.002 |
|  | ε-FTSVR | 0.202 ± 0.102 | 0.972 ± 0.152 | 0.325 ± 0.142 | 0.009 |
|  | ε-TSVR | 0.213 ± 0.103 | 0.960 ± 0.156 | 0.333 ± 0.143 | 0.011 |
| Auto Price | ε-HFTSVR | 0.296 ± 0.072 | 0.945 ± 0.242 | 0.369 ± 0.069 | 0.0011 |
|  | ε-FTSVR | 0.325 ± 0.077 | 0.935 ± 0.237 | 0.375 ± 0.086 | 0.0015 |
|  | ε-TSVR | 0.334 ± 0.083 | 0.924 ± 0.235 | 0.393 ± 0.093 | 0.002 |

TABLE III
BEST PARAMETERS OF ε-HFTSVR AND ε-FTSVR ON UCI DATASETS

| Regressor: ε-HFTSVR | | | |
|---|---|---|---|
| Dataset | $p_1 = p_2$ | $p_3 = p_4$ | $\varepsilon_1 = \varepsilon_2$ |
| Servo | 0.0032 | 0.0032 | 0.0064 |
| Auto Price | 8 | 0.3152 | 0.14 |
| Regressor: ε-FTSVR | | | |
| Servo | 0.0036 | 0.0036 | 0.0072 |
| Auto Price | 4 | 0.3142 | 0.22 |

REFERENCES

[1] C. Burges, "A Tutorial on Support Vector Machines for Pattern Recognition," Data Mining and Knowledge Discovery, vol. 2, no. 2, pp. 121–167, 1998.
[2] A. Chaudhuri and K. De, "Fuzzy Support Vector Machine for Bankruptcy Prediction," Applied Soft Computing, vol. 11, no. 2, pp. 2472–2486, 2011.
[3] N. Y. Deng, Y. J. Tian and C. H. Zhang, *Support Vector Machines: Theory, Algorithms and Extensions.* CRC Press, 2012.
[4] Jayadeva, R. Khemchandani and S. Chandra, "Twin Support Vector Machines for Pattern Classification," IEEE Transactions on Pattern Analysis and Machine Intelligence, vol. 29, no. 5, pp. 905–910.
[5] A. Samola and B. Schöikopf, "A Tutorial on Support Vector Regression," Statistics and Computing, vol. 14, pp. 199–222, 2004.
[6] Y. H. Shao, C. H. Zhang, Z. M. Yang, L. Jing and N. Y. Deng, "An ε-Twin Support Vector Machine for Regression," Neural Computing and Applications, vol. 23, no.1, pp. 175–185, 2012.
[7] H. J. Zimmermann, *Fuzzy Set Theory and its Applications*. Boston: Kluwer Academic, 2001.
[8] X. Peng, "TSVR: An efficient Twin Support Vector Machine for Regression," Neural Networks, vol. 23, no. 3, pp. 365–372, 2010.
[9] UCI Datasets: http://archive.ics.uci.edu/ml/